\UseRawInputEncoding
\documentclass[journal,10pt,sin ]{IEEEtran}
\usepackage{amsmath,amsfonts}
\usepackage{algorithmic}
\usepackage{array}
\usepackage[caption=false,font=normalsize,labelfont=sf,textfont=sf]{subfig}
\usepackage{textcomp}
\usepackage{stfloats}
\usepackage{url}
\usepackage{verbatim}
\usepackage{graphicx}
\usepackage{xcolor}
\usepackage{bm}
\def\BibTeX{{\rm B\kern-.05em{\sc i\kern-.025em b}\kern-.08em
    T\kern-.1667em\lower.7ex\hbox{E}\kern-.125emX}}
\usepackage{balance}
\usepackage{subfig}

\begin{document}
\title{UNDERSTANDING URBAN WATER CONSUMPTION USING REMOTELY SENSED DATA}

\author{\IEEEauthorblockN{ Shaswat Mohanty\IEEEauthorrefmark{1}\IEEEauthorrefmark{2}, Anirudh Vijay\IEEEauthorrefmark{1}\IEEEauthorrefmark{3}, Shailesh Deshpande\IEEEauthorrefmark{4} \\}
\IEEEauthorblockA{\IEEEauthorrefmark{1}Equal contribution \\}
\IEEEauthorblockA{\IEEEauthorrefmark{2}Department of Mechanical Engineering, Stanford University, CA \\}
\IEEEauthorblockA{\IEEEauthorrefmark{3}Department of Electrical Engineering, Stanford University, CA \\}
\IEEEauthorblockA{\IEEEauthorrefmark{4}Tata Research Development and Design Centre}
\vspace{-3em}
}

\maketitle

\begin{abstract}
Urban metabolism is an active field of research that deals with the estimation of emissions and resource consumption from urban regions. The analysis could be carried out through a manual survey or by the implementation of elegant machine learning algorithms. In this exploratory work, we estimate the water consumption by the buildings in the region captured by satellite imagery. To this end, we break our analysis into three parts: i) Identification of building pixels, given a satellite image, followed by ii) identification of the building type (residential/non-residential) from the building pixels and finally iii) using the building pixels along with their type to estimate the water consumption using the average per unit area consumption for different building types as obtained from municipal surveys.
\end{abstract}

\vspace{-1em}
\section{INTRODUCTION}
\IEEEPARstart{U}{rban} metabolism models cities as living entities that consume energy and natural resources, releasing byproducts in the process. While some of these could be beneficial and create economic value, the others are undesired such as emission and pollution \cite{kennedy2011study}. Sustainable planning incorporates urban metabolism to optimize the consumption of resources while limiting those byproducts. 

Understanding the behavior of the entire system is facilitated by modeling the behavior of its constituent subsystems. Remote sensing technology provides a synoptic view of the region over a long period of time and is useful in estimating the parameters of urban metabolism. Satellite data provides insights into important characteristics of cities that can be used to extract features for classification of the subsystems , especially hyperspectral data is found to be more useful in quantifying urban metabolism~\cite{camps2005kernel}. The classification itself can be done using traditional machine learning algorithms such as K-means clustering and support vector machines or deep learning algorithms. The results of the classification can be analyzed to find the correlation between features extracted from images to resource consumption. To estimate water consumption in a given area, for example, the number and type of buildings would be useful features. Related approaches to identify the effect of other subsystems on urban metabolism such as roofing material and urban growth~\cite{samsudin2016development,yadav2018computational}, building identification~\cite{cohen2016rapid}, road identification~\cite{saito2015building} and vegetation identification~\cite{stojanova2010estimating}, have also been carried out using machine learning approaches.

In this paper, we discuss the process of preparing the training data comprising multispectral satellite images and non-linear topographical features for our learning algorithm. The study features a comparison between three learning algorithms for the identification and classification of building pixels. We proceed to show how we leverage this information along with resource consumption survey data to estimate local resource consumption from multispectral satellite images.
\vspace{-1em}
\section{METHODOLOGY} \label{sec:method}
\noindent  Our process flow for this project can be broadly classified as follows: i) Input: An eight-band satellite image of a region. ii) Process 1: Identifying the type of building and how similar proxies can be used for this classification. iii) Process 2: Manipulate datasets to incorporate additional topographical information and correlate them to emissions from a region. iv) Output: Building identification, classification, and resource consumption/emission estimation of the satellite image.
Building classification is vital in being able to estimate the resource consumption. The multi-step approaches can be broken into: i) Classification of building and non-building pixels (restricted to pixel-level analysis; not extending to pixel clustering). ii) Classification of residential and non-residential building pixels from the identified building pixels. iii) Water consumption estimation of the region captured in the image.
The following portions of this section discuss the approach in detail.
\vspace{-1em}
\subsection{Dataset}
\noindent We are using the Defence Science and Technology Laboratory (DSTL) Satellite Imagery Feature Detection dataset\cite{kaggle}. The satellite image dataset comes annotated with labels for buildings whose position and shape data are in the form of polygons. The polygons are defined as sets of ordered coordinates marking the corners of the buildings. The dataset contains 3-band images (red, blue, and green channels) and 8-band multispectral images (red, red edge, coastal, blue, green, yellow, near-IR1, and near-IR2 channels). The 3-band images have a resolution of 0.31 m and the multispectral images have a resolution of 1.24 m. We consider the tasks of detecting buildings and classifying them as residential or non-residential and provide an estimate of water consumption in the region. For the numerical results presented in this paper, we prepare our dataset by expanding the feature space. The details of the feature space expansion are discussed later in this section.
\vspace{-1em}
\subsection{Classification Algorithms}

\noindent In this section we discuss the three algorithms we test in our paper - stochastic gradient descent, random forests classifier, and multilayer perceptron neural network. The training set size is large ($\sim$ 11 million vectors), and so, SVM and other kernel-based methods are not feasible. We test our models against a baseline model, which is trained without feature space expansion, i.e, using only the RGB values of every pixel. We briefly describe the methods used for classification problem in this subsection. 

The Stochastic Gradient Descent (SGD) algorithm is widely used for machine learning problems due to ease of defining a baseline model. The SGD algorithm is classified by the loss function and regularization used. In our study, we test the accuracy obtained by using the \textit{logistic} and \textit{modified\_huber} loss function. 

Random forests (RF) are a kind of ensemble learning technique that use multiple decision trees and perform classification by combining their results \cite{liaw2002classification}. For classification, the final prediction is taken to be the mode of the individual predictions of the trees in the forest. Each of the trees is trained by choosing a random subset of the training data. The number of trees, the degree of branching, and the size of the random subset are among the different hyperparameters in the learning process, that we sample from to arrive at a locally optimal classification model. A successful segmentation based approach carried out by~\cite{bialas2019optimal} gives us reason to believe that our pixel frame width parameter to expand the feature space might help us in obtaining reliable results.

Multilayer perceptrons (MLP) are a type of feed-forward ANNs with multiple layers of nodes and nonlinear activations. Some of the popular activation functions are \textit{ReLU}, \textit{tanh} and \textit{sigmoid}. The more popular implementation of this algorithm is for other aspects of urban metabolism such as the identification of damaged buildings~\cite{tesfamariam2010earthquake} and gasoline source classification (resource consumption/source identification) using multispectral data~\cite{balabin2008gasoline}.
\vspace{-1em}
\subsection{Building Detection and Expansion of Feature Space}

\noindent The workflow in analyzing the image is shown in Fig.~\ref{fig1}(a) for brevity. We generate a raster ground-truth image from the training dataset. It indicates building pixels as white pixels (positive class) and non-building pixels as black pixels (negative class). After generating the ground truth, we enhance the feature space to increase the amount of information fed into the classifier and to provide additional context to each pixel. The complexity of the expansion of the feature space can be controlled by the pixel frame width parameter, $k$, as shown in Fig.~\ref{fig1}(a) where we consider the neighboring pixels from $k$ pixel thick square. This results in the expansion of the feature space $\in \!R^{c(2k+1)^{2}}$, where $c$ is the number of color channels for each pixel ($c=8$ for the hyperspectral data we are analysing in this paper). 

Additionally, the oriented gradients around every pixel obtained from our histogram of oriented gradients descriptor add non-linear complexity to our feature space. The Histogram of Oriented Gradients (HOG) is a popular feature descriptor used for object detection and image processing \cite{dalal2005histograms}. Typically, the HOG is computed locally for different blocks of pixels. For machine learning using images, this technique helps augment the feature space with spatial information.
\vspace{-1em}
\subsection{Metrics for Deciding Optimal Model}
\noindent To evaluate the accuracy and precision, we are targeting the following metrics\footnote{ i) True positive ($TP$), ii) True negative ($TN$), iii) False positive ($FP$), iv) False negative ($FN$)}: 
\begin{enumerate}
    \item Pixel Jaccard, $P_{J}=\frac{TP}{TP+FN+FP}$,
    \item Positive class accuracy, $A_{p}=\frac{TP}{TP+FN}$,
    \item Negative class accuracy, $A_{n}=\frac{TN}{FP+TN}$,
    \item Balanced accuracy, $\bar{A}=\frac{A_p+A_n}{2}$, and,
    \item AUC (area under curve) which is the area under the ROC (receiver operating characteristic) curve.
\end{enumerate}
During prediction, we threshold the predicted probabilities at the optimal threshold value that maximizes the Pixel Jaccard. While we perform k-fold cross-validation later to obtain quantitative performance metrics, we break the image into two halves to visualize the results.
\vspace{-1em}
\subsection{Residential vs Non-Residential Classification}
\noindent Classifying building pixels according to building type can help us approximate emissions and resource consumption from an area captured by a satellite image. Resource consumption by buildings is a function of the type of buildings. Residential and commercial buildings require different quantities of resources. We make the assumption that residential and non-residential buildings have different emission/consumption rates per unit area and that they remain constant across all the buildings of one type. To avoid interpreting the feature extracted by unsupervised learning and to utilize the model we obtain from building pixel identification, we approached this problem as a supervised learning problem. Upon visual inspection, we identified possible clusters of non-residential buildings and manually labelled the dataset for this classification. 

\begin{figure*}[ht]
    \centering
    \subfloat[a]{\includegraphics[width=0.22\textwidth]{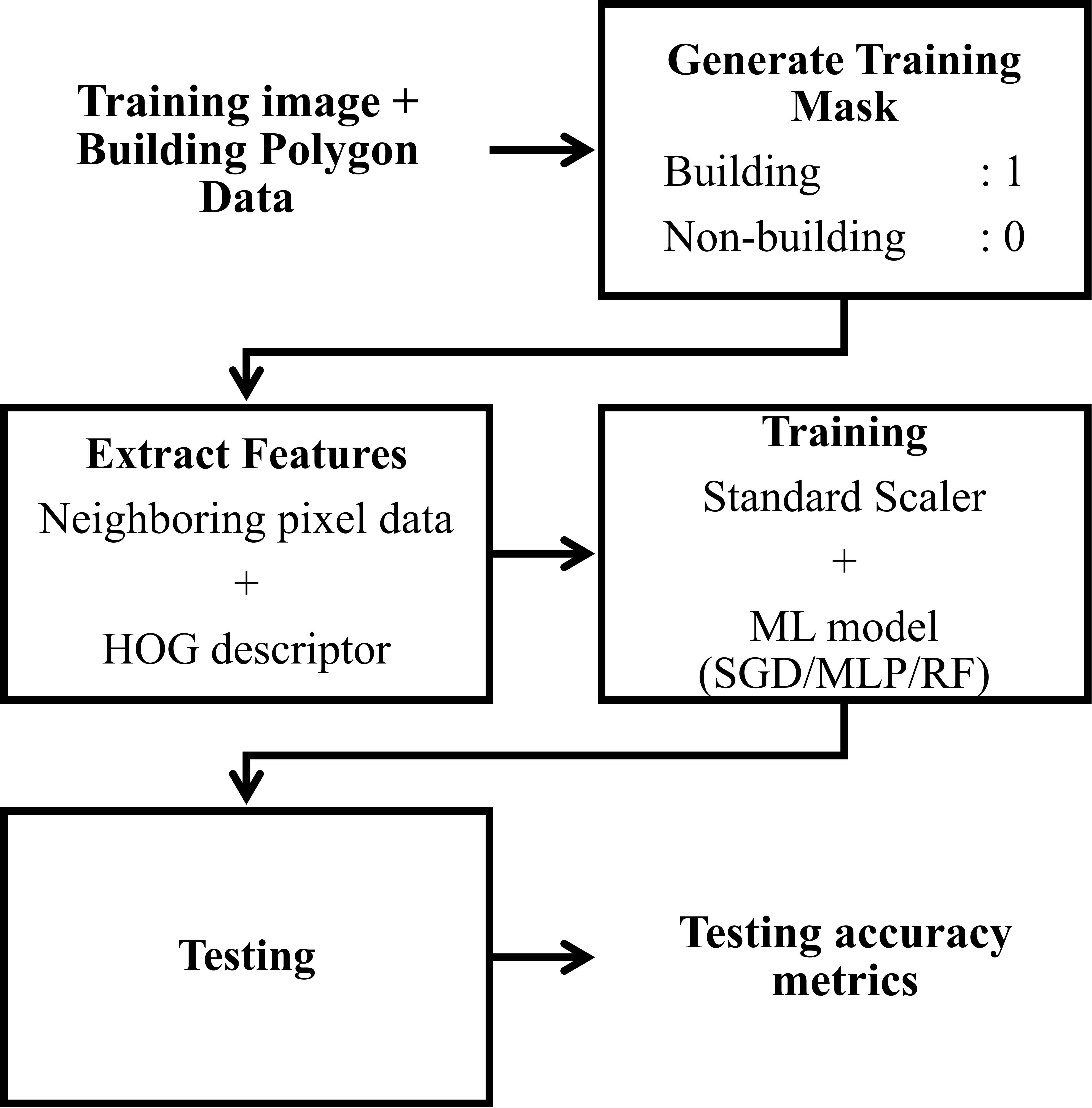}}
    \subfloat[b]{\includegraphics[width=0.31\textwidth]{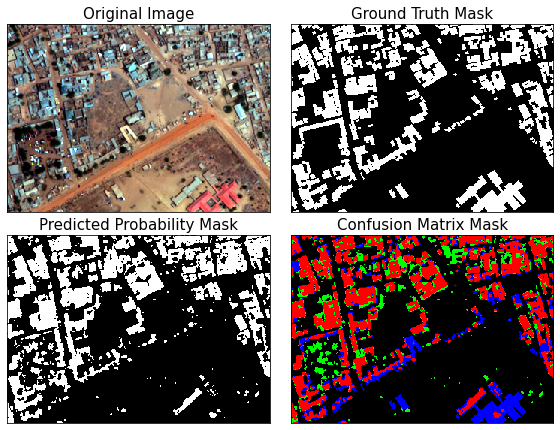}}
    \subfloat[c]{\includegraphics[width=0.33\textwidth]{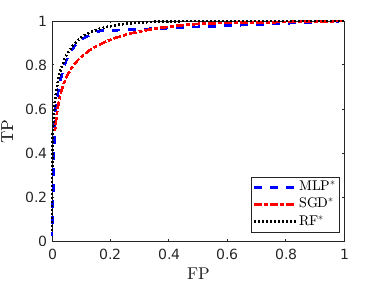}}
    \caption{(a) A schematic describing our baseline workflow. (b) Top left: The rendered RGB subset of the test image; Top right: Ground truth mask. High-intensity pixels indicate buildings and low-intensity pixels indicate non-buildings; Bottom left: Predicted probability mask. The intensity is represents the probability of the pixel belonging to the building class; Bottom right: Confusion matrix mask obtained after thresholding the predicted probability mask and comparing it with the ground truth. \textcolor{red}{Red}, \textcolor{green}{green}, and \textcolor{blue}{blue} pixels indicate \textcolor{red}{TP}, \textcolor{green}{FP}, and \textcolor{blue}{FN}, respectively. (c) ROC curve for the SGD, RF, and MLP Classifiers indicating that the RF Classifier is marginally better.}     \label{fig1}
    \vspace{-2em}
\end{figure*}
\vspace{-1em}

\section{RESULTS} \label{sec:results}
\noindent The AUC mentioned Table.~\ref{tab:my_label} can be observed in ROC curve shown in Fig.~\ref{fig1}(c). The computation of the AUC is done by the trapezoidal method and can be found in the MATLAB script in our zipped folder submission.

\vspace{-1em}
\subsection{Building Pixel Classification}
\noindent Fig.\ref{fig1}(a) shows the working of the building classification and table.~\ref{tab:my_label} shows the training accuracies of logistic regression (base-line) and other conventional ML algorithms. To identify the optimal model for building classification, we tuned the hyper-parameters for SGD, MLP, and RF Classifiers. The metric to decide the optimal set of parameters was the optimum $\bm{P_{J}}$. Our findings can be summarized as i) The SGD Classifier results in the lowest $\bm{P_{J}}$ and has a threshold $>0.5$, ii) The RF and MLP Classifier have comparable performances, that are superior to the SGD classifier, iii) RF performs marginally better than the MLP classifier, as shown in Fig.~\ref{fig1}(c), however, the threshold for the optimum $\bm{P_{J}}$ is closest to $0.5$ for the MLP Classifier and iv) the training accuracy is indicative that RF is susceptible to overfitting. Therefore, we have finalized the optimal MLP Classifier as the ideal model for our analysis. These results have been tabulated in Table.~\ref{tab:my_label}. 
\begin{table}[ht]
\caption{Optimal performance metrics ($\alpha$ and $\beta$ denote testing and training values, respectively).}
\begin{tabular}{|c|c|c|c|c|c|c|c|}
    \hline
       \textbf{Model} & $\bm{\bar{A}}^{\alpha}$ &  $\bm{P_{J}}^{\alpha}$ & $\bm{AUC}^{\alpha}$ &  $\bm{\bar{A}}^{\beta}$ &$\bm{P_{J}}^{\beta}$ &  $\bm{\bar{A}}^{\beta}$ &  $\bm{P_{J}}^{\beta}$\\
        \hline
      Base   & 0.722 & 0.532 & 0.843 &0.856 &0.602& 0.896 & 0.598 \\ 
     MLP$^{*}$ & 0.898& 0.730& 0.952 & 0.961&0.883& 0.999& 0.994\\  
     RF$^{*}$ & 0.903& 0.740& 0.974 & 0.999& 0.996& 1.0& 0.999\\ 
      SGD$^{*}$ & 0.858& 0.656& 0.939 &0.863 &0.664& 0.903& 0.638\\ 
    \hline
    \end{tabular}
    \label{tab:my_label}
\end{table}
\vspace{-1em}
For all the models under consideration we carried out hyper-parameter tuning by considering a limited space of hyper-parameters and $<M>^{*}$ refers to the optimal set of hyper-parameters for model $<M>$. The accuracy metric $\langle . \rangle$ for the residential and non-residential building is denoted by $\langle \bar{.} \rangle$. The optimal hyper-parameter set for the three classifiers is as follows: 
\begin{enumerate}
     \item SGD Classifier: \texttt{loss='log', alpha=1e-3, class\_weight='balanced', threshold = 0.62}
    \item RF Classifier: \texttt{n\_estimators=500, max\_depth=50, min\_samples\_leaf=2, min\_samples\_split=2, class\_weight='balanced', threshold = 0.39}
   
    \item MLP Classifier: \texttt{hidden\_layer\_sizes=(75, 25, 100, 20, 75, 25), max\_iter=1000, threshold = 0.46}
\end{enumerate}

To extract the test accuracy of our model, we have carried out a $5$-fold cross-validation and reported the mean optimum accuracy obtained for every fold.

\vspace{-1em}
\subsection{Residential vs Non-residential Building Classification} \label{resnonres}
\noindent The building type was annotated manually using visual interpretation metrics, e.g., large building size, roofing materials, proximity to building clusters, roads, and industrial structures like storage tanks, etc. The mask is rendered for all pixels for visualization purposes. However, only the vectors corresponding to the building pixels are fed into the classifier. We formulate this as a binary classification problem with residential buildings as the positive class and non-residential buildings as negative class. In the mask (Fig.~\ref{fig1}(b)), the residential buildings are shown in gray (training)/red (prediction) and the non-residential buildings are shown in white (training)/yellow (prediction).  

The features that are used in this problem are the same features that were used for the building pixel detection problem except the features generated by the HOG descriptor.
For this classification problem, we run our base-line model without the HOG descriptor. 
\vspace{-1em}
\begin{figure}[h]
    \centering
    \subfloat{\label{fig:scheme}\includegraphics[width=0.35\textwidth]{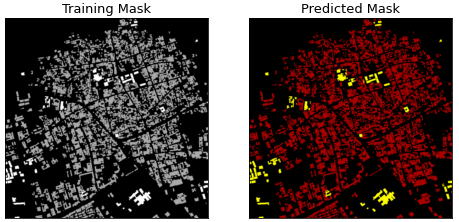}}\label{workfaslow}
    \caption{The ground truth and the prediction mask where the residential buildings are shown in gray (training)/red (prediction) and the non-residential buildings are shown in white (training)/yellow (prediction). } \label{masks4bldg}
\end{figure}
We then compare it against the other classifiers for the optimal hyper-parameter space, for $k=4$. The training accuracy metrics for the classification models are shown in Table.~\ref{tab:my_label}. We observe that the RF Classifier is still susceptible to overfitting for this problem. The analysis could be improved by carrying out a similar hyper-parameter tuning for the different classifiers over the validation set to obtain a more reliable model for residential vs non-residential building classification. 
\subsection{Resource Consumption/Emission Estimation}

\noindent To estimate the water consumption in buildings in the image, we first estimate the area covered by residential buildings ($A_R$) and the area covered by non-residential buildings ($A_{NR}$). This consumption is estimated by,
\vspace{-0.5em}
\begin{align*}
    A_R = \sum_{\text{pixel}~i} &( a_P\times \mathbf{P}\{i\in \text{B} \}
    &\times \mathbf{P}\{i\in \text{R}| i\in \text{B}\} ),
\end{align*}
\vspace{-1.5em}
\begin{align*}
    A_{NR} = \sum_{\text{pixel}~i} &( a_P\times \mathbf{P}\{i\in \text{B} \}
    &\times \mathbf{P}\{i\in \text{NR}| i\in \text{B}\} ),
    \vspace{-1.5em}
\end{align*}
where, $a_P$ is the area of one pixel in $m^2$, B, R and NR, denote building, residential and non-residential, respectively. From the predicted mask obtained from our analysis in Section.~\ref{resnonres}, as shown in Fig.~\ref{masks4bldg}, we computed the areas as, $A_R=213858$ $m^2$, and $A_{NR}=16988$ $m^2$. This figure is then weighted by the consumption per unit area available in \cite{Senn:2009} \footnote{The water usage statistics is not well documented for different building types and for all locations across the globe. Therefore, we have used the data for USA, which was readily available} to calculate the water consumption ($W$),
\vspace{-0.5em}
\begin{equation*}
    W = A_R\times W_R + A_{NR}\times W_{NR},
    \vspace{-0.5em}
\end{equation*}
where, $W_R$ and $W_{NR}$ are the water-consumption values per unit area for residential and non-residential buildings, respectively. As per \cite{Senn:2009},  $W_R=40$ ${\rm gal/person}$, and $W_{NR}=21$ ${\rm gal/person}$. These values are obtained by averaging the per sq-ft values for commercial and residential buildings in \cite{Senn:2009}. Using this information, and the assumption that for either building-type an individual occupies $750$ ft$^2$, we can obtain the daily water consumption from the $1$ $\text{km}^2$ image at $\approx 0.128$ million gallons, where residential buildings account for $\approx 0.123$ million gallons, whereas non-residential buildings account for the remaining $\approx 0.005$ million gallons of daily water consumption. Since, the satellite images are not geotagged, we cannot validate our results against a know region. To this end, we compare our results against the daily water usage per km$^2$ usage of similar landscapes in USA such as Phoenix ($\approx 0.194$ million gallons) and Portland ($\approx 0.091$ million gallons). Our prediction is within $40\%$ of the reported values for similarly landscaped cities.
\vspace{-1em}
\section{CONCLUSION} \label{sec:conclusion}
\noindent Through our analysis of a few classification algorithms and parameter tuning of their hyper-parameter space, we were able to identify the Multi-Layer Perceptron Classifier as the optimal classifier for the building vs non-building pixel detection problem, with a training accuracy of $0.961$ and a test accuracy of $0.894$. This model was less prone to overfitting than the Random Forest Classifier.

The computational efficiency of our residential vs. non-residential classification was decreased by the manual preparation of the dataset. In the expansive urban metabolism problem, identifying the type of buildings is the intermediary step between building pixel identification and emission estimation. We could obtain an optimal model with a training accuracy of $\approx 0.999$, however, this could be an artifact of over-fitting, possibly due to the small data set, that further analysis could help explain.

Given that water consumption data was readily available \cite{Senn:2009}, we could use a simple closed-form expression that utilizes the result of our building type classification and the data from \cite{Senn:2009} to give an approximation of the daily water consumption from the buildings observed in the image spanning $1$ $\text{km}^2$. Furthermore, the process described in this paper is designed for third-party inspecting and verification of the urban footprint on resource consumption. It can generate dynamic and frequent updates to the estimates of demand and consumption surpassing the need for exhaustive surveys.

In this paper, we successfully demonstrate building pixel classification, and our breakdown of the future scope discusses the improvements that we could make to achieve our secondary goals (building type classification and emission/resource consumption estimation).

\nopagebreak
\vspace{-1em}

\end{document}